# Anomaly Detection for Multivariate Time Series on Large-scale Fluid Handling Plant Using Two-stage Autoencoder

Susumu Naito
*Corporate R&D Center*
*Toshiba Corporation*
Kawasaki, Japan
susumu.naito@toshiba.co.jp

Yasunori Taguchi
*Corporate R&D Center*
*Toshiba Corporation*
Kawasaki, Japan
yasunori.taguchi@toshiba.co.jp

Kouta Nakata
*Corporate R&D Center*
*Toshiba Corporation*
Kawasaki, Japan
kouta.nakata@toshiba.co.jp

Yuichi Kato
*Corporate R&D Center*
*Toshiba Corporation*
Kawasaki, Japan
yuichi12.kato@toshiba.co.jp

*Abstract*— This paper focuses on anomaly detection for multivariate time series data in large-scale fluid handling plants with dynamic components, such as power generation, water treatment, and chemical plants, where signals from various physical phenomena are observed simultaneously. In these plants, the need for anomaly detection techniques is increasing in order to reduce the cost of operation and maintenance, in view of a decline in the number of skilled engineers and a shortage of manpower. However, considering the complex behavior of high-dimensional signals and the demand for interpretability, the techniques constitute a major challenge. We introduce a Two-Stage AutoEncoder (TSAE) as an anomaly detection method suitable for such plants. This is a simple autoencoder architecture that makes anomaly detection more interpretable and more accurate, in which based on the premise that plant signals can be separated into two behaviors that have almost no correlation with each other, the signals are separated into long-term and short-term components in a stepwise manner, and the two components are trained independently to improve the inference capability for normal signals. Through experiments on two publicly available datasets of water treatment systems, we have confirmed the high detection performance, the validity of the premise, and that the model behavior was as intended, i.e., the technical effectiveness of TSAE.

I. INTRODUCTION

Multivariate time series data are becoming ubiquitous in real-world applications. For example, in a power plant, the complex interplay of the numerous factors constituting the plant status is measured with thousands of sensors for intelligent operation, monitoring, and maintenance. Or take another example: in the field of semiconductor manufacturing, hundreds of equipment monitoring sensors check equipment properties that affect quality and productivity. The use of multivariate time series data has attracted much attention because of its potential to solve problems in various fields.

Anomaly detection for multivariate time series data is essential for application. In the real world, there is a growing need for anomaly detection techniques to reduce the cost of operations and maintenance, in view of the decline in the number of skilled engineers and the shortage of manpower. As for the power plant example, by detecting early signs of anomalies based on the complex relations among signals from multiple sensors, the safety, efficiency, and reliability of the plant can be greatly improved. Automated, high-performance anomaly detection techniques are in high demand in various fields.

Much research has been done on detecting unexpected behavior for anomaly detection in multivariate time series [1]. The most commonly used techniques include probabilistic approaches such as Hotelling's $T^2$ [2], distance approaches such as k-nearest neighbors [3], and classification approaches such as One-Class SVM [4]. However, for example, it is difficult to use them to comprehensively monitor the signals of a power plant, which change in a complex way even under normal conditions, because the number of signals is too large. As the number of dimensions increases, the performance of these techniques generally deteriorates because of the curse of dimensionality.

Most recently, unsupervised anomaly detection methods based on deep learning, which infer correlations between time series and identify anomalous behaviors, have been attracting attention [5][6][7][8][9]. Deep learning-based methods provide high performance with complex network architecture. One of the problems with these methods is the interpretability of the modeling. Since users of anomaly detection applications are primarily plant experts, in order to trust the anomaly detection results, they first expect the modeling to be understandable in direct relation to their knowledge of the system and data. Insufficient interpretability makes it difficult for them to accept the method, even if it has a high performance.

In this work, we focus on the case of anomaly detection in a large-scale plant, specifically a fluid handling plant with dynamic components, such as a power generation, water treatment, or chemical plant, where signals from various physical phenomena are observed simultaneously. In a power plant, one dominant signal is a 'long-term' fluctuation, where large amplitude changes occur in most of the sensors over the long term, for example, accompanied by plant-scale operations and maintenance. Another dominant signal is a 'short-term' fluctuation, where relatively small amplitude changes occur in a portion of sensors in the short term, for example, accompanied by pump oscillation and local thermal changes. This feature is assumed to be true for other fluid handling plants. By explicitly considering these different types of fluctuation, anomaly detection is expected to be more interpretable and more accurate.

For interpretable anomaly detection modeling of a large-scale plant, we introduce a simple deep learning-based algorithm Two-Stage AutoEncoder (TSAE) [10]. We extend a preliminary case study of anomaly detection on a nuclear power plant with TSAE [10], and make clear that TSAE modeling is interpretable and preferable for multivariate time series data on plants. Remarkably, although TSAE has rather simple network architecture, it performs better than state-of-the-art methods in anomaly detection for publicly available datasets of water treatment systems.

The main contributions of this paper are:

- We introduce TSAE as an anomaly detection method suitable for various large-scale fluid handling systems.
- We perform an empirical study on publicly available datasets to analyze the technical effectiveness of TSAE by comparing the performance of TSAE with those of the state-of-the-art methods and assessing the effect of the TSAE architecture.
- We conduct an intensive investigation on parameters and clarification of pre-processing settings for reproducibility. Compared to image data, pre-processing of multivariate time series data is not well established. In particular, we show details of normalization that sometimes causes problems for anomaly detection in time series data when train and test sets are normalized separately.
- Through analysis of the experimental results, we show an inherent problem in anomaly detection: identity mapping problem. Too high expressiveness can reconstruct anomalies, leading to the miss detection of anomalies.

The rest of this document is organized as follows. Section II discusses methods for detecting unsupervised anomalies in multivariate time series. Section III discusses the details of our method. Sections IV and V describe the experimental setup including the pre-processing and the experiments on our method, respectively.

II. RELATED WORK

Anomaly detection for multivariate time series is an active topic. Supervised learning methods [11][12] are able to identify only known anomaly types [13]. For example, in power plants, when an anomaly occurs, measures are taken to prevent it from reoccurring. Therefore, most of the anomalies that will occur in the future are of unknown types. The use of supervised methods is limited and unsupervised approaches are effective. Many of the unsupervised methods for multivariate time series anomaly detection in the literature use autoencoders or recurrent neural networks. The state-of-the-art methods can be categorized into the following types:

- Autoencoders [7][14][15]. UnSupervised Anomaly Detection for multivariate time series (USAD) is a method based on an autoencoder architecture [7] whose learning is inspired by Generating Adversary Networks (GAN) [16]. The intuition behind USAD is that the adversarial training of its encoder-decoder architecture allows it to learn how to amplify the reconstruction error of inputs containing anomalies, while gaining stability compared to methods based on GANs architectures.
- Recurrent neural networks [9][17][18]. OmniAnomaly [9], a stochastic recurrent neural network for multivariate time series anomaly detection, learns robust multivariate time series' representations with a stochastic variable connection and a planar normalizing flow, and uses the reconstruction probabilities to determine anomalies.
- Hybrids [8][19][20]. LSTM-VAE [8] combines a Long Short-Term Memory (LSTM) with a Variational AutoEncoder (VAE) by replacing the feed-forward network in a VAE with an LSTM. The LSTM-VAE models the time dependence of time series through LSTM networks and obtains a better generalization capability than traditional methods.

Our study provides a new perspective on these types of state-of-the-art methods by constructing an architecture that reflects the general characteristics of multivariate time series data on systems with a common physical mechanism. This perspective can also contribute to the performance improvement of the methods reported in these previous studies.

III. METHOD

*A. Problem Formulation*

We denote multivariate time series as a sequence of data points:

$$\mathcal{T} = \{x_1, \ldots, x_T\}, x \in \mathbb{R}^m. \qquad (1)$$

Each one is an observation of a process measured at a specific time *t*. Multivariate time series contain *m* variables at each time instant. We focus on an unsupervised learning problem where $\mathcal{T}$ is given as training input. $\mathcal{T}$ contains only normal points. Anomaly detection is to identify an unseen observation $\hat{x}_t$, $t > T$, where it is different from $\mathcal{T}$. The amount of difference between the unseen sample $\hat{x}_t$ and the normal dataset $\mathcal{T}$ is measured by an anomaly score, which is then compared to a threshold to obtain an anomaly label. In order to formulate the dependence between a current time point and previous ones, we define $W_t$ as a time window of length *K* at given time *t* :

$$W_t = \{x_{t-K+1}, \ldots, x_{t-1}, x_t\}. \qquad (2)$$

The original time series $\mathcal{T}$ can be transformed into a sequence of windows $\mathcal{W} = \{W_1, \ldots, W_T\}$ to be used as training input. We define a label $y_t$, $y \in \{0,1\}$, to indicate a detected anomaly at time *t* , i.e. $y_t = 1$, or not ($y_t = 0$). The goal of our anomaly detection problem is to assign the label $y_t$ to $\hat{x}_t$ in an unseen window $\widehat{W}_t, t > T$, based on the anomaly score. That is, a current time point and previous ones are used to infer the current time point, and an anomaly is identified with the current time point.

## B. Two-stage Autoencoder

An Autoencoder (AE) [20] is an unsupervised artificial neural network combining an encoder network and a decoder network. It sets the output of the decoder to be equal to the input of the encoder. The encoder network learns a compressed representation of the input. The decoder network reconstructs the input of the encoder from the compressed representation. The difference between the input $X$ and the reconstruction $AE(X)$ is called the reconstruction error. The autoencoder learns so as to minimize the reconstruction error $\mathcal{L}_{AE}$ as an objective function. It is defined as:

$$\mathcal{L}_{AE} = \|X - AE(X)\|^2, \quad (4)$$

where $\|\cdot\|$ is the L2-norm.

Figure 1 shows a configuration of TSAE. TSAE consists of two autoencoders of $AE_1$ and $AE_2$. $AE_1$ and $AE_2$ are designed to learn long-term and short-term components of signals, respectively. How each autoencoder works for multivariate time series data is stated in the next section. In order to obtain the reconstruction $R$ of TSAE, $AE_1$ and $AE_2$ are connected as follows:

$$W'_t = AE_1(W_t), \quad (5)$$

where $W'_t = \{x'_{t-K+1}, \dots, x'_{t-1}, x'_t\}$.

$$dx_t = x_t - x'_t, \quad (6)$$

$$dx'_t = AE_2(dx_t), \quad (7)$$

$$R_t = x'_t + dx'_t. \quad (8)$$

The input of TSAE is a time window and the output of TSAE is the reconstruction of a time instant. The reconstruction error of TSAE is the anomaly score.

The training process is summarized in Algorithm 1. This consists of two steps. First, $AE_1$ learns so as to minimize the reconstruction error $\mathcal{L}_{AE_1}$ of $AE_1$.

$$\mathcal{L}_{AE_1} = \|W_t - W'_t\|^2. \quad (9)$$

Next, $AE_2$ learns so as to minimize the reconstruction error $\mathcal{L}_{AE_2}$ of $AE_2$.

$$\mathcal{L}_{AE_2} = \|dx_t - dx'_t\|^2. \quad (10)$$

The detection process is summarized in Algorithm 2. The anomaly score is defined as:

$$\mathcal{A}(\hat{x}_t) = \|\hat{x}_t - \hat{R}_t\|^2. \quad (11)$$

If the anomaly score $\mathcal{A}(\hat{x}_t)$ is higher than a defined anomaly threshold, the $\hat{x}_t$ is declared abnormal.

## C. Properties of TSAE

Here, we show that the TSAE architecture is interpretable and preferable for multivariate time series data in fluid handling systems.

In real plants, various physical phenomena occur in mutually related facilities, and sensors observe the signals of these overlapping phenomena. Based on the experts' knowledge of power plants, we assume that the complex

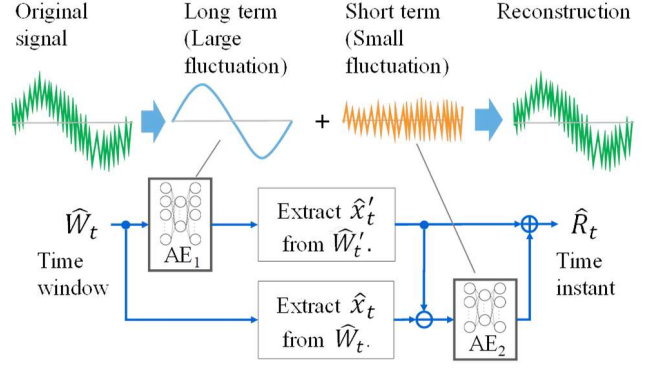

Fig.1. TSAE Configuration

---

**Algorithm 1 TSAE training algorithm**

**STEP 1.**

**Input:** Normal windows Dataset $\mathcal{W} = \{W_1, \dots, W_T\}$, Number of epochs $N_1$

**Output:** Trained $AE_1$

  $AE_1 \leftarrow$ initialize weights
  **for** $n = 1$ to $N_1$ **do**
    **for** $t = 1$ to $T$ **do**
      $W'_t \leftarrow AE_1(W_t)$
      $\mathcal{L}_{AE_1} \leftarrow \|W_t - W'_t\|^2$
      $AE_1 \leftarrow$ update weights using $\mathcal{L}_{AE_1}$
    **end for**
  **end for**

**STEP 2.**

**Input:** Trained $AE_1$, Normal windows Dataset $\mathcal{W} = \{W_1, \dots, W_T\}$, where $W_t = \{x_{t-K+1}, \dots, x_{t-1}, x_t\}$, Number of epochs $N_2$

**Output:** Trained $AE_2$

  $AE_2 \leftarrow$ initialize weights
  $n \leftarrow 1$
  **for** $n = 1$ to $N_2$ **do**
    **for** $t = 1$ to $T$ **do**
      $W'_t \leftarrow AE_1(W_t)$, where $W'_t : \{x'_{t-K+1}, \dots, x'_{t-1}, x'_t\}$
      $dx_t \leftarrow x_t - x'_t$
      $dx'_t \leftarrow AE_2(dx_t)$
      $\mathcal{L}_{AE_2} \leftarrow \|dx_t - dx'_t\|^2$
      $AE_2 \leftarrow$ update weights using $\mathcal{L}_{AE_2}$
    **end for**
  **end for**

---

behavior of the process signals can be separated into two types: long term and short term.

*Long term.* Large amplitude fluctuations in the signals due to changes in the plant operating conditions, such as power variation, normal operation, maintenance, etc. In the long term,

```
Algorithm 2 TSAE detection algorithm
```

**Input:** Test windows Dataset $\widehat{\mathcal{W}} = \{\widehat{W}_1, \ldots, \widehat{W}_{T^*}\}$, where $\widehat{W}_t = \{\hat{x}_{t-K+1}, \ldots, \hat{x}_{t-1}, \hat{x}_t\}$, threshold $\lambda$

**Output:** Labels $\mathcal{Y}$: $\{y_1, \ldots, y_{T^*}\}$

    **for** $t=1$ to $T^*$ **do**

        $\widehat{W}_t' \leftarrow AE_1(\widehat{W}_t)$, where $\widehat{W}_t': \{\hat{x}'_{t-K+1}, \ldots, \hat{x}'_{t-1}, \hat{x}'_t\}$

        $\widehat{dx_t} \leftarrow \hat{x}_t - \hat{x}'_t$

        $\widehat{dx'_t} \leftarrow AE_2(\widehat{dx_t})$,

        $R_t \leftarrow \hat{x}'_t + \widehat{dx'_t}$.

        $\mathcal{A}(\hat{x}_t) \leftarrow \|\hat{x}_t - \hat{R}_t\|^2$

        **if** $\mathcal{A}(\hat{x}_t) > \lambda$ **then**

            $y_t \leftarrow 1$

        **else**

            $y_t \leftarrow 0$

        **end if**

    **end for**

a large number of the signals change simultaneously (global correlation).

*Short term.* Small amplitude fluctuations in the signals due to small vibrations of dynamic components, such as pump pulsation (flow oscillation), oscillation of heat transfer rate in a heat exchanger due to flow oscillation (temperature oscillation), etc. They are almost random in the time direction but have correlations between the signals at each time instant. In the short term, a small number of the signals change simultaneously (local correlation).

In power plants, there is little correlation between these two behaviors. It is assumed that this uncorrelation is also valid for fluid handling systems.

We can expect to reconstruct the normal state with high accuracy by learning these uncorrelated two behaviors independently. The premise of TSAE application is that the signal can be separated into two almost uncorrelated behaviors. TSAE is designed to learn the long-term and short-term components of the signal respectively by combining two different types of autoencoders, $AE_1$ and $AE_2$. As illustrated in Fig1, $AE_1$ is a network to learn and infer only large fluctuations (the long-term components). Because the contribution of the small fluctuations (the short-term components) to the reconstruction error is small, $AE_1$ is set to be a less expressive network in order to ignore their contribution as negligible in learning. $AE_1$ is typically a shallow network with one intermediate layer. To learn temporal correlation in the long term, the time window, that is, signals of a certain period of time is set as the input and output of the $AE_1$. The small fluctuations, the input of $AE_2$, are extracted as the difference between the input and output of $AE_1$. In $AE_2$, in order to learn and infer the short-term components, which are almost random in the time direction but have correlations between signals, signals of a single time frame are set as the input and output of $AE_2$. With these interpretable structures, we consider that TSAE actively learns different kinds of signals in plants, and provides an explanation for its high performance as shown below.

## IV. EXPERIMENTAL SETUP

This section describes the datasets, pre-processing, hyper-parameters, implementation, and the performance metrics used in the experiments.

### A. Public Datasets

Two publicly available datasets of water treatment systems were used in our experiments. Table I summarizes the characteristics of datasets. The excluded signals are explained in the next section. The datasets are briefly described in the following.

*Secure Water Treatment (SWaT) Dataset.* The SWaT dataset[1] is a scaled-down version of an actual industrial water treatment plant producing filtered water [22]. The collected dataset [23] consists of 11 days of continuous operation: 7 days collected under normal operations and 4 days collected with 41 attack scenarios. The dataset has two versions: Version 0, which includes the first 30 minutes of the plant start-up operation, and Version 1, which excludes the first 30 minutes and includes only normal operation; we used Version 1 because the attacks are performed only during normal operation.

*Water Distribution (WADI) Dataset.* This dataset[2] is collected from the WADI testbed, an extension of the SWaT testbed [23]. It consists of 16 days of continuous operation, of which 14 days were collected under normal operation and 2 days with 15 attack scenarios. As the plant was unstable for certain periods during the operation, we used an updated version "WADI_14days_new.csv" from which the affected readings have been removed.

### B. Pre-processing

*Normalization.* The training data were normalized by scaling the range (the minimum value, the maximum value) of each signal in the training data to the range (0, 1). For many signals, the training data and the test data have different ranges in the same signal. Therefore, the test data were normalized using the same scaling factors as the training data. We emphasize that we used the same scaling factors in both the training and test data. Unlike the image data, multivariate time series data on many plants including SWaT and WADI, even for normal signals have different ranges in the training and test data. If the training and test data are independently normalized by min-max scaling, the model outputs misguided inference results. In [7], we suppose that this is one of the main reasons for the small F1 score in the WADI dataset.

*Excluded signals.* The signals in Table I (2B_AIT_002_PV for WADI, AIT201 and P201 for SWaT) were excluded from the training and test data. As shown in Fig. 2, these signals in the test data deviate significantly from the ranges in the training data, regardless of the times of the ground truth anomaly segments. These deviations, even if normal, should be detected as anomalies by the anomaly detection methods, because they are behaviors that are not present in the training data. We excluded them because they lead to incorrect performance evaluation of the method.

---

[1] https://itrust.sutd.edu.sg/itrust-labs_datasets/dataset_info/#swat
[2] https://itrust.sutd.edu.sg/itrust-labs_datasets/dataset_info/#wadi

TABLE I. Benchmarked Datasets. (%) is the percentage of anomalous data points in the dataset.

| Dataset | Train | Test | Dimensions | Anomalies (%) | Excluded signals |
|---|---|---|---|---|---|
| WADI | 784571 | 172801 | 122 | 5.77 | 2B_AIT_002 |
| SWaT | 495000 | 449919 | 49 | 12.14 | AIT201, P201 |

*Down-sampling.* To reduce computational cost, down-sampling was performed for each dataset by applying an anti-aliasing filter and then resampling it. The down-sampling process was performed using scipy.signal.decimate[3] with its default setting. For all experiments except parameter effects in Sec. V.E, the down-sampling rate was set to 5 for all datasets, as in [7].

### C. TSAE Hyper-parameters and Implementaion

In our experiments, we set hyper-parameters of TSAE to empirical values or to the same values as in [7]. $AE_1$ and $AE_2$ of TSAE were fully connected autoencoders with one intermediate layer. The activation functions in the intermediate and output layers were sigmoid functions. The number of dimensions of the intermediate layer of $AE_1$ and that of $AE_2$ were 1/2 of that of the input layer of $AE_1$ and 1/10 of that of the input layer of $AE_2$, respectively. The window size $K$ was the same as in [7], 10 for WADI, and 12 for SWaT. The training data were randomly divided into training : validation = 4 : 1. Step size for sliding the window was set to 1. The training epochs were 200 for $AE_1$ and 20 for $AE_2$ in all experiments. $AE_2$ converges in a smaller number of epochs than $AE_1$ because of its smaller dimensionality. The packages and versions used for the algorithm implementation were: python 3.7.3, tensorflow 1.14.0, scikit-learn 0.20.2, numpy 1.19.2, and scipy 1.1.0. As an optimizer, we used Adam operator with its learning rate of 0.0001 for $AE_1$ and 0.001 for $AE_2$. All experiments are conducted on NVIDIA GeForce GTX 1080 Ti 11GB GDDR5X GPU. Using the above hyper-parameters, the training times of our model for WADI and SWaT were about 29 and 9 minutes respectively.

### D. Performance Metrics

We used Precision (P), Recall (R), and F1 score (F1) to evaluate anomaly detection performance:

$$P = \frac{TP}{TP + FP}, \qquad R = \frac{TP}{TP + FN}, \qquad F1 = 2 \cdot \frac{P \cdot R}{P + R},$$

where TP, FP, FN are the numbers of True Positives, False Positives, and False Negatives.

Performance is assessed by comparing the results of each evaluated method with the annotated ground truth. In our performance calculations, we adopted the "point-adjust" approach proposed by [8] and used in [7][9]. In practice, anomalous observations usually occur continuously to form contiguous anomaly segments. In this approach, if any observation in the ground truth anomaly segment is detected correctly, all observations in the segment are considered to have been correctly detected as anomalies. The observations outside the ground truth anomaly segment are treated as usual. In actual power plant anomaly detection, the number of times the anomaly score exceeds the threshold for an anomaly is of little significance, and what is important is whether the anomaly was detected or not. The same can be assumed for water treatment plants, chemical plants, etc. Therefore, we adopted point-adjust as a realistic approach.

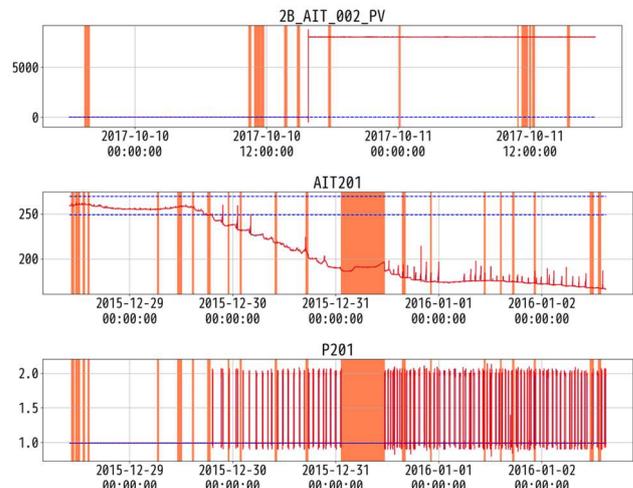

Fig. 2. Excluded signals (2B_AIT_002_PV for WADI, AIT201 and P201 for SWaT). Red lines show the signals. Orange vertical lines show the times of the ground truth anomaly segments. Blue dashed lines show the maximum and minimum values of the signals in the training data. For 2B_AIT_002PV and P201, the maximum and minimum vales are almost the same

### V. EXPERIMENTS AND RESULTS

The technical effectiveness of TSAE was assessed by the following two procedures. First, we assessed the presence or absence of a performance superiority of TSAE by comparing its performance with performances of the state-of-the-art methods. Secondly, we assessed whether or not the performance of TSAE is due to the effect of TSAE architecture by verifying the assumption that the signals can be separated into two uncorrelated components and by verifying whether TSAE behaved as we intended. In addition, we analyzed the performances of the state-of-the-art methods, investigated the substitutability of the TSAE architecture with multi-layered AEs, and investigated the parameter effects of TSAE.

### A. Overall Performance

To show the overall performance of TSAE, we compared it with four unsupervised methods for multivariate time series anomaly detection. These are AE [21] as a baseline method and LSTM-VAE [8], USAD [7], and OA [9] as the state-of-the-art methods. AE has the same configuration and implementation as $AE_1$ in TSAE in Sec. IV. C. In the experiment, we used publicly available codes for LSTM-VAE[4] and OA[5]. For USAD, we implemented it based on [7]. Here, in LSTM-VAE and USAD, each anomaly score is calculated in time window units ($\widehat{W}_t$), whereas in TSAE and OA, each anomaly score is calculated in time instant units ($\hat{x}_t$). For their comparison, AE was evaluated for both the anomaly score in time window units (AE-w), $\|\widehat{W}_t - \widehat{W}_t'\|^2$, and that in time instant units (AE-i), $\|\hat{x}_t - \hat{x}'_t\|^2$, since in our pre-experiments AE showed significant difference between them. We tested the possible thresholds for all models and recorded the results with the highest F1 score. All experiments were performed three times for each method to reduce the effect of random seeds, and their average values were reported as the final results. Table II details the performance results obtained for all the methods on the public datasets. The standard deviation of F1 score was 0.001 or less for AE-w, AE-i, and TSAE, 0.02 for LSTM-VAE and USAD, and 0.01 for OA in WADI, and 0.01 or less for all methods in SWAT. Table II shows that TSAE has a higher F1 score than other methods in WADI, confirming its performance superiority. In SWaT, the average F1 score of all methods was as high as 0.874. The difference between the maximum and minimum F1 scores was as small as 0.048. There was almost no difference in performance between the methods.

---
[3]https://docs.scipy.org/doc/scipy/reference/generated/scipy.signal.decimate.html
[4]https://github.com/Danyleb/Variational-Lstm-Autoencoder
[5]https://github.com/NetManAIOps/OmniAnomaly

TABLE II. Performance comparison. Precision (P), recall (R) and F1 score in WADI and SWaT datasets.

| Methods | WADI | | | SWaT | | |
|---|---|---|---|---|---|---|
| | *P* | *R* | *F1* | *P* | *R* | *F1* |
| AE-w | 0.591 | 0.640 | 0.615 | 0.876 | 0.861 | 0.868 |
| AE-i | 0.532 | **1.000** | 0.695 | 0.874 | **0.930** | **0.901** |
| LSTM-VAE | 0.465 | 0.694 | 0.543 | **0.883** | 0.825 | 0.853 |
| USAD | 0.484 | 0.722 | 0.575 | 0.874 | 0.843 | 0.858 |
| OmniAnomaly | 0.540 | 0.858 | 0.663 | 0.877 | 0.852 | 0.864 |
| TSAE | **0.635** | **1.000** | **0.777** | 0.871 | **0.930** | 0.899 |

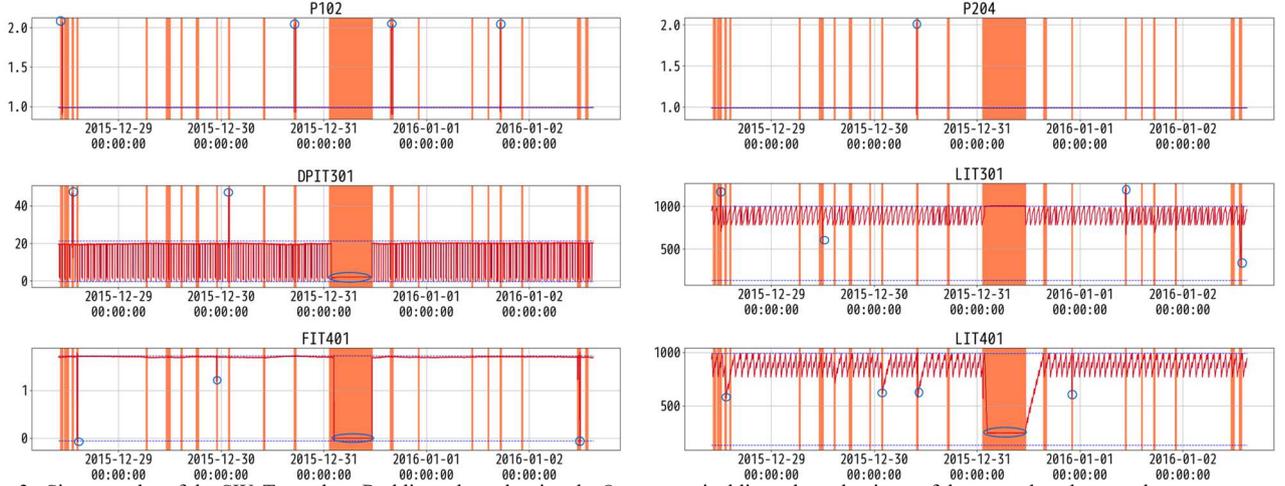

Fig. 3. Six examples of the SWaT test data. Red lines show the signals. Orange vertical lines show the times of the ground truth anomaly segments. Blue dashed lines show the maximum and minimum values of the signals in the training data. Blue circles show change points of the signals at the times of the ground truth anomaly segments.

Figure 3 shows some examples of the SWaT test data. In SWaT, one or more signals change significantly at the times of the ground truth anomaly segments, which made it easy to detect the anomalies. As a result, the F1 scores became high and the differences among the methods were less apparent. Therefore, in order to clarify the characteristics of TSAE, subsequent descriptions focused on WADI, where there is a clear difference in F1 scores among the methods.

### B. Effect of the TSAE Architecture

In order to confirm the effect of the TSAE architecture on the performance results in the previous section, we first verified that the signals can be separated into two uncorrelated components, which is a premise for the TSAE application. Figure 4 shows a heat map of the correlation coefficients of the outputs ($x'_t$) of AE$_1$ and the deviations ($dx_t$) from the inputs (the training data of AE$_2$), in the training data of the WADI dataset. Figure 4 indicates that the outputs of AE$_1$ and the deviations are almost uncorrelated, confirming that the signals can be separated into two uncorrelated components. This satisfies the premise for TSAE, and the WADI dataset is suitable data for TSAE.

In addition, we verified whether the outputs ($x'_t$) of AE$_1$ and the deviations ($dx_t$) have global and local correlations, respectively, as we assume in Sec III.C. Figure 5 shows the results of hierarchically clustering the correlation coefficients of the outputs of AE$_1$ and the deviations. From Fig. 5, the outputs of AE$_1$ have a global correlation (a large number of signals changing simultaneously), and the deviations have a local correlation (a small number of signals changing simultaneously). This suggests that our assumption is a general property of fluid handling systems.

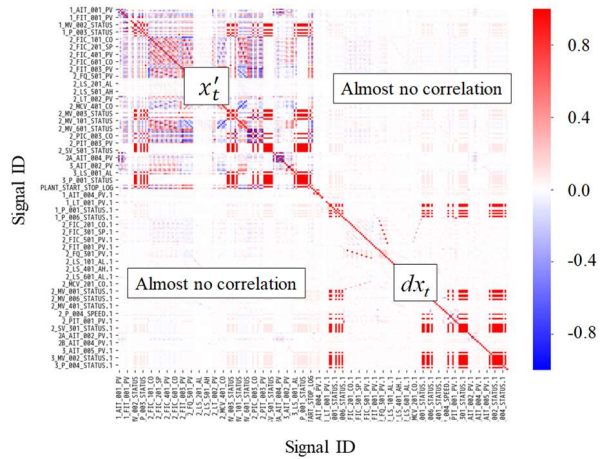

Fig. 4. Heat map of the correlation coefficients of the output of AE$_1$ ($x'_t$), and the deviation ($dx_t$) in the training data on WADI dataset. The labels on the vertical and horizontal axes are first all 122 signals for $x'_t$ and then all 122 signals for $dx_t$ in the same order as $x'_t$.

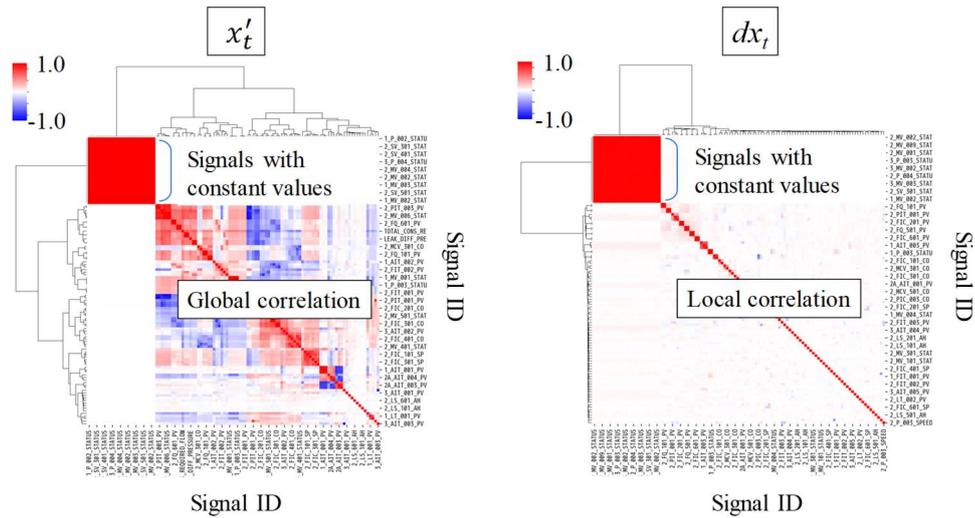

Fig. 5. The results of hierarchical clustering of the correlation coefficients of AE$_1$ ($x'_t$), and the deviation ($dx_t$) in the training data on WADI dataset. The left and right figures are heat map for $x'_t$ and that for $dx_t$, respectively.

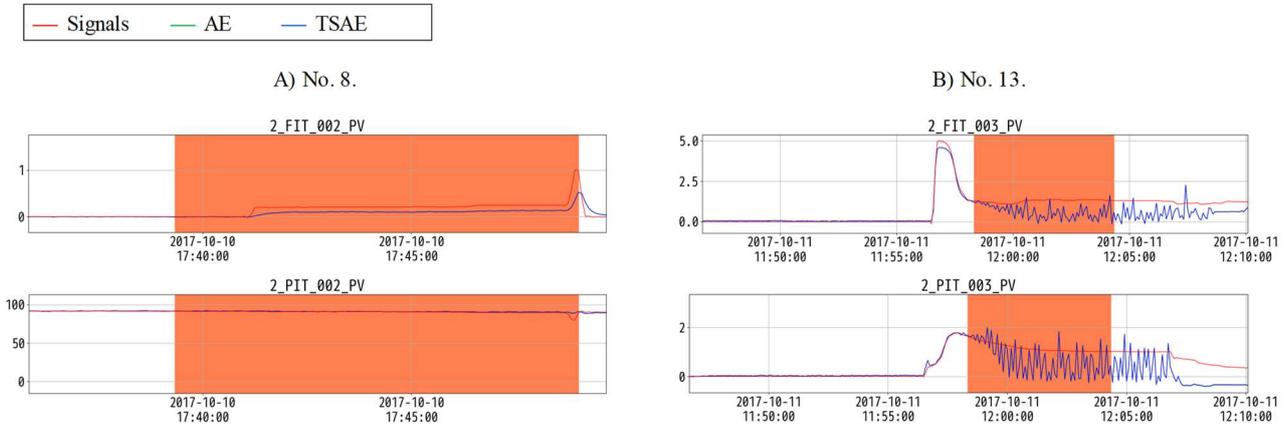

Fig. 6. The trend graphs of A) No. 8 and B) No. 13 attacks. Red, green, and blue lines show the signals that the attacks should reflect, the outputs of AE, and those of the TSAE, respectively. Orange vertical lines show the times of the ground truth anomaly segments.

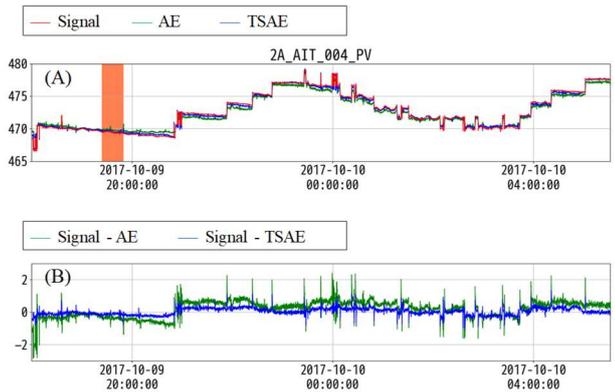

Fig. 7. Example of A) a trend graph for a normal signal, the outputs of TSAE and AE, and B) the differences between the normal signal and the outputs of TSAE and AE. In (A), red, green, and blue lines show the normal signal, the output of AE, and that of TSAE, respectively. Orange vertical lines show the times of the ground truth anomaly segments. In (b), green and blue lines show the difference between the normal signal and the output of AE, and that between the normal signal and the output of TSAE, respectively.

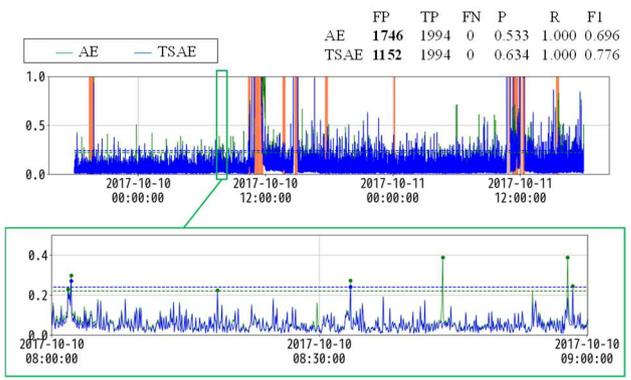

Fig. 8. Trend graph of the anomaly scores of AE and TSAE (the upper figure), and its enlarged graph (the lower figure). Green and blue lines show the anomaly scores in AE and in TSAE, respectively. Green and blue dashed lines show the thresholds at the best F1 score in AE and in TSAE, respectively. In Fig. 7, their false positives (FP), true positives (TP), precision (P), recall (R), and F1 score are also shown.

Next, we verified whether TSAE behaved as we intended by comparing TSAE with AE-i. Regarding the behavior of TSAE, our intention is for the outputs of TSAE to have better agreement with the normal signals by reconstructing the short-term fluctuations of the normal signals, and in the case of the anomaly signals, for the outputs of TSAE to behave almost the same as those of AE if the anomalies are long-term fluctuations.

TABLE III. Performance comparison of AEs and TSAEs. The number of intermediate layers, precision (P), recall (R) and F1 score in WADI dataset, network configuration, the number of nodes and the number of edges. The configuration denotes a ratio of the number of nodes in each layer to the number of dimensions in the input layer.

| Methods | Mid layers | P | R | F1 | Configurations | Nodes | Edges |
|---|---|---|---|---|---|---|---|
| AE-i | 1 | 0.532 | 1.000 | 0.695 | 1→1/2→1 | 1,830 | 1,488,400 |
| AE-i | 3 | 0.642 | 0.857 | 0.732 | 1→1/2→1/4→1/2→1 | 2,745 | 1,860,500 |
| AE-i | 5 | 0.681 | 0.791 | 0.732 | 1→1/2→1/4→1/8→1/4→1/2→1 | 3,202 | 1,953,525 |
| TSAE | 1 | 0.635 | 1.000 | 0.777 | $AE_1$: 1→1/2→1 | 2,013 | 1,502,064 |
| TSAE | 3 | 0.677 | 0.979 | 0.800 | $AE_1$: 1→1/2→1/4→1/2→1 | 2,928 | 1,874,164 |
| TSAE | 5 | 0.714 | 0.791 | 0.750 | $AE_1$: 1→1/2→1/4→1/8→1/4→1/2→1 | 3,385 | 1,967,189 |

We first compared their behaviors with respect to the anomalies. The explanation sheet of the WADI dataset[2] shows which signals the attack should reflect, for 3 out of all 15 attacks. In one of these three attacks, there was no difference between the anomaly signal and their reconstruction results for both TSAE and AE. In the remaining two attacks there were differences between the anomaly signals and them. Figure 6 shows the trend graphs of the two attacks (No. 8 and No. 13). In Fig. 6, the red lines (the signals) in the anomaly segments change gradually, not instantaneously, so the anomalies are long-term fluctuations. The green lines (AE) almost coincide with the blue lines (TSAE), and thus the green lines (AE) are not visible on the graphs. This indicates that the behaviors of TSAE with respect to the anomalies were almost the same as those of AE, which is consistent with the behaviors that we intended. In addition, in Fig. 6 (B), the curves of AE and TSAE deviate from the observations even after the time of the anomaly segment because the anomaly segment shows only the time of the attack, and the transient state until the subsequent return to the normal state is not included in the segment.

We then compared the behaviors of TSAE and AE with respect to the normal signals. Figure 7 shows an example of a trend graph for a normal signal and the differences between the normal signal and the outputs of TSAE and AE. Figure 7 indicates that for the normal signal, the spikes in the difference for TSAE are smaller than those for AE and the difference for TSAE approaches zero on average, i.e., the output of TSAE has better agreement with the normal signal. This behavior agrees with the behavior that we intended. Figure 8 shows a trend graph of the anomaly scores of TSAE and AE. As a result of the improved reconstruction of TSAE with respect to the normal signals, the spikes in the anomaly score of TSAE at the times of the normal segments are smaller than those of AE. In Fig. 8, FP, TP, P, R, and F1 are also shown. There is no difference in the TP between TSAE and AE, reflecting that the behaviors of TSAE with respect to the anomaly signals are almost the same as those of AE. On the other hand, FP of TSAE is smaller than that of AE, reflecting that the reconstruction of the short-term components (spikes) of the normal signals was improved and the spikes in the anomaly score for the normal signals became smaller. The F1 score increased due to the smaller FP, although the TP did not change. TSAE behaved as we intended, and its effect resulted in the highest F1 score.

In conclusion, the signals can be separated into two uncorrelated components, and TSAE behaved as we intended.

Therefore, the high-performance result of TSAE is due to the effect of the TSAE architecture.

### C. Analysys of the State-of-the-art Methods

In this section, we analyze the performance results of the state-of-the-art methods in the WADI dataset in Table II by comparing them with that of AE, the baseline method. LSTM-VAE and USAD, whose anomaly scores are calculated in time window units, were compared with AE-w. OA, whose anomaly score is in time instant units, was compared with AE-i. Finally, AE-w and AE-i were compared.

In Table II, LSTM-VAE and USAD have lower precision than AE-w, resulting in lower F1 scores. For each signal, LSTM-VAE outputs a curve with its change suppressed as the reconstruction, which is similar to what is evident in [8], and was inferior to AE in reconstructing fast changes in the signals. This resulted in more variability of anomaly score at the times of normal segments, more false positives, and lower precision. In USAD, the objective function is designed, roughly speaking, to perform normal training (minimization of ||normal training data - model output||$_2$) for a few epochs after the start of training, and then perform training to improve discrimination performance with pseudo-generated data. For the WADI dataset, the normal training was insufficient compared to 200 epochs in AE-w, and the reconstruction of fast changes in normal signals was insufficient compared to AE, which resulted in more false positives and lower precision due to more variability of anomaly score at the times of normal segments. These results indicate that in order to reduce the number of false positives, it is important to adequately reconstruct normal signals.

In Table II, OA has a slightly lower F1 score than AE-i by 0.03. OA has higher precision and lower recall than AE-i. This is because OA has fewer false positives and more false negatives than AE-i. We consider that this result indicates identity mapping problem. In anomaly detection, the reconstruction result of a model is required to be consistent with normal signals (FP reduction) and different from anomaly signals (FN reduction). The higher expressiveness of a model than complexity of data has a risk that the model is trained to output the input as is. If this occurs in anomaly detection, the reconstruction result agrees with the anomaly signals and the anomalies cannot be detected. We call this phenomenon identity mapping problem in anomaly detection. The performance result of OA shows that OA has a tendency to identity mapping compared to AE-i. In terms of identity mapping problem, this result indicates that a simple model can be more effective than a complex model with high expressiveness.

Finally, we compare the performance results of AE-w and AE-i. The F1 score of AE-i in time instant units was higher than that of AE-w in time window units. AE-i had lower precision than AE-w because of more false positives, and AE-i had higher recall than AE-w because of fewer false negatives. The reconstruction error (anomaly score) in time window units is averaged over time within a time window. The rapid change in the reconstruction error is suppressed in time window units, compared to that in time instant units. Therefore, false positives due to fast changes in normal signals such as spikes are less likely to occur in time window units, whereas fast changes in anomaly signals are more difficult to detect. Due to this trade-off, for the WADI dataset, the anomaly score in time instant units was more effective. Naturally, better reconstruction of fast changes in normal signals leads to better detection with time instant units assumed to retain higher recall. We consider TSAE to be a model that satisfies this property.

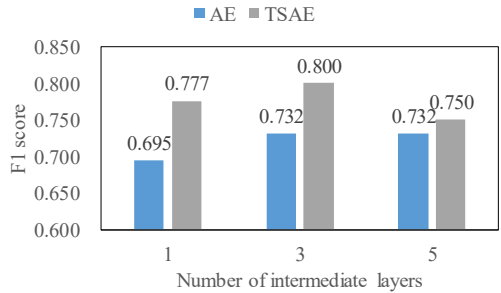

Fig. 9. F1 scores of AEs and TSAEs with different number of intermediate layers. Blue and gray bars show the F1 scores in AEs and those in TSAEs, respectively. Horizontal labels show the number of intermediate layers of AE for the AE and that of $AE_1$ for TSAE.

### D. Substitutability of the TSAE Architecture

In Secs. V. A and B, the technical effectiveness of TSAE was confirmed. However, a simple question remains, "Didn't the F1 score go up simply because there were more nodes and layers than in AE?". We investigated this question. Table III and Fig. 9 show the performance of AEs with different numbers of intermediate layers and that of TSAEs. In all experiments, $AE_2$ of TSAE had one intermediate layer. In addition to the number of intermediate layers, Table III shows the number of nodes and edges as a measure of the model's expressiveness. The F1 score of AE is improved by increasing the number of intermediate layers. However, the F1 score of TSAE with one intermediate layer of $AE_1$, which has fewer hyper-parameters and thus less expressiveness, was higher than that of AE with three or five intermediate layers. This indicates that the F1 score is improved mainly by the effect of the TSAE architecture, the two-stage configuration based on the characteristics of the actual data, not by the increase in the number of hyper-parameters.

From Table III, the recall decreases as the number of intermediate layers increases for both AE and TSAE. This result indicates a tendency to identity mapping and in anomaly detection, simply increasing the expressiveness does not necessarily result in good performance.

### E. Effect of Parameters

In this section, we investigate the various parameters and factors that can impact on the performance of TSAE. All experiments were conducted using the WADI dataset.

The first factor we investigate is how TSAE responds to various down-sampling rates of the training data. Down-sampling reduces the computational cost but can have a negative effect by causing a loss of information. Figure 10 summarizes the results obtained using five different rates [1, 5, 10, 20, 50]. The results show that the F1 score is almost invariant at rates 1 and 5, but decreases as the rate further increases. Down-sampling is less desirable in TSAE.

The next factor we investigate is how TSAE responds to various window sizes in the data. The window size has an impact on the anomaly detection performance, as a larger window size can capture slower rising and falling signals. Figure 10 summarizes the results obtained using five different window sizes [5, 10, 20, 50, 100]. The results show a gradual increase in the F1 score as the size increases; for TSAE, the larger the window size, the better the performance.

The final factor we investigate is how TSAE responds to the variation in the number of nodes in the model. A small number of nodes is insufficient for learning features, whereas too many nodes can cause identity mapping problem, making it difficult to detect anomalies. Figure 10 summarizes the results obtained using six different combinations of the number of nodes. The results show that the F1 score is almost insensitive to the number of nodes. Compared to the results in Sec. V. D, we speculate that the number of intermediate layers is one, which strongly limits the change in the performance of the model.

In summary, for TSAE, down-sampling is less desirable, larger window size is better, and it is almost insensitive to the number of nodes. The results of down-sampling and window size indicate that better performance can be obtained through learning more features by expanding the window size without losing information by down-sampling. The results of the number of nodes indicates the robustness of TSAE.

### VI. CONCLUSIONS

We introduced the Two-Stage AutoEncoder (TSAE) as an anomaly detection method for various fluid handling systems with dynamic components, such as power generation, water treatment, and chemical plants. This is a simple autoencoder

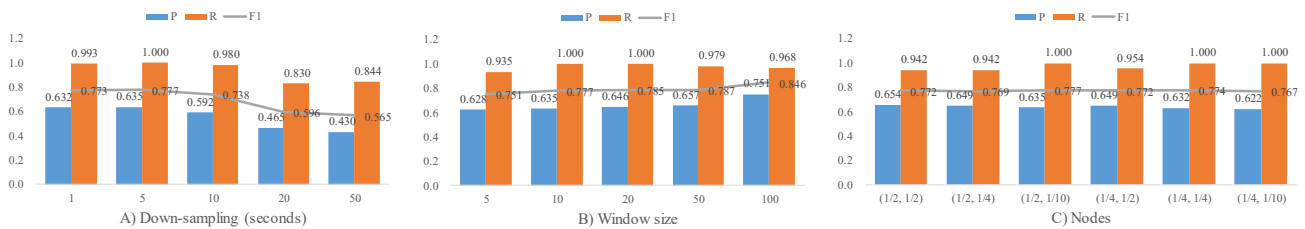

Fig. 10. Effect of parameters. Precision (P), Recall (R) and F1 score (F1) as a function of A) the down-sampling rate, B) the window size $K$, C) combinations of the number of nodes $(a, b)$, where $a$ and $b$ show a ratio of the number of nodes in the intermediate layer of $AE_1$ to that of dimensions in the input layer of $AE_1$, and a ratio of the number of nodes in the intermediate layer of $AE_2$ to that of dimensions in the input layer of $AE_2$

architecture that make anomaly detection more interpretable and more accurate, in which based on the premise that plant signals can be separated into two behaviors that have almost no correlation with each other, the signals are separated into long-term and short-term components in a stepwise manner, and the two components are trained independently to improve the inference capability for normal signals.

To verify the technical effectiveness of TSAE, we conducted an empirical study using publicly available datasets of water treatment systems. In the WADI dataset, where performance differences occurred between methods, TSAE obtained a higher F1 score than other state-of-the-art methods, confirming the performance superiority of TSAE. Since 1) the signals could be separated into two behaviors that are almost uncorrelated with each other, which is a premise for TSAE, and 2) TSAE behaved as we intended, improving the ability to infer the normal state and thus both detecting anomalies and reducing false positives, we confirmed that the high F1 score is due to the effect of the TSAE architecture. With these results, we could confirm the technical effectiveness of TSAE in the WADI dataset. Because of the technical effectiveness in WADI, which consists of pumps and piping similar to a power plant, we infer that TSAE is also effective in other fluid handling systems with dynamic components.

From the analysis of the results of the state-of-the-art methods and the investigation of the substitutability of the TSAE architecture, we could identify a problem specific to anomaly detection. Although it is important to reconstruct fine fluctuations in normal signals in order to reduce false positives, identity mapping is a problem inherent in anomaly detection, and good performance cannot be obtained by simply increasing expressiveness of a model. From the viewpoint of preventing identity mapping, a simple model such as TSAE can be more effective than complex models with high expressiveness, as in the case of the WADI dataset. Our experiments on parameter effects show that for TSAE, learning more features without losing information from the training data results in a better F1 score. We consider this to be a reasonable result indicating that the performance is better when learning without losing information.